\documentclass[10pt,twocolumn,letterpaper]{article}

\usepackage[pagenumbers]{cvpr} 

\usepackage{graphicx}
\usepackage{amsmath}
\usepackage{amssymb}
\usepackage{booktabs}
\usepackage{lipsum}
\usepackage[table,xcdraw]{xcolor}
\usepackage{multirow}
\usepackage{tikz}

\usepackage{algorithm}
\usepackage{algorithmic}
\usepackage{comment}
\usepackage{xcolor}
\usepackage{ifthen}
\newboolean{commentsVisible}
\setboolean{commentsVisible}{true}

\ifthenelse{\boolean{commentsVisible}}{
    \newcommand{\FR}[1]{\textcolor{magenta}{FR$>$ #1}}
    \newcommand{\JY}[1]{\textcolor{cyan}{JY$>$ #1}}
    \newcommand{\old}[1]{\textcolor{gray}{OLD$>$ #1}}
    \newcommand{\todo}[1]{\textcolor{purple}{(ToDo: #1)}}

}{
    \newcommand{\FR}[1]{}
    \newcommand{\JY}[1]{}
    \newcommand{\old}[1]{}
    \newcommand{\todo}[1]{}
}

\definecolor{mymagenta}{HTML}{9673A6}
\definecolor{mygreen}{HTML}{4D8101}

\usepackage[pagebackref,breaklinks,colorlinks]{hyperref}

\usepackage[capitalize]{cleveref}
\crefname{section}{Sec.}{Secs.}
\Crefname{section}{Section}{Sections}
\Crefname{table}{Table}{Tables}
\crefname{table}{Tab.}{Tabs.}

\def\cvprPaperID{9140} 
\def\confName{CVPR}
\def\confYear{2024}

\begin{document}

\title{CaLDiff: Camera Localization in NeRF via Pose Diffusion}

\author{%
Rashik Shrestha$^{1}$ \quad Bishad Koju$^{1}$ \quad Abhigyan Bhusal$^1$ \quad Danda Pani Paudel$^{1,2}$ \quad François Rameau$^{1,3}$\\
$^1$NepAl Applied Mathematics and Informatics Institute for research (NAAMII) \\ \quad $^2$ETH Zürich \quad $^3$State University of New York (SUNY) at Korea \\
\texttt{\{rashik.shrestha,bishad.koju,abhigyan.bhusal\}@naamii.org.np}\\
\texttt{paudel@vision.ee.ethz.ch, francois.rameau@sunykorea.ac.kr}
\vspace{-1em}
}


\maketitle

\begin{abstract}
With the widespread use of NeRF-based implicit 3D representation, the need for camera localization in the same representation becomes manifestly apparent. Doing so not only simplifies the localization process -- by avoiding an outside-the-NeRF-based localization -- but also has the potential to offer the benefit of enhanced localization. This paper studies the problem of localizing cameras in NeRF using a diffusion model for camera pose adjustment. More specifically, given a pre-trained NeRF model, we train a diffusion model that iteratively updates randomly initialized camera poses, conditioned upon the image to be localized. At test time, a new camera is localized in two steps: first, coarse localization using the proposed pose diffusion process, followed by local refinement steps of a pose inversion process in NeRF. In fact, the proposed camera localization by pose diffusion (CaLDiff) method also integrates the pose inversion steps within the diffusion process. Such integration offers significantly better localization, thanks to our downstream refinement-aware diffusion process. Our exhaustive experiments on challenging real-world data validate our method by providing significantly better results than the compared methods and the established baselines. Our source code will be made publicly available.
  
\end{abstract}

\section{Introduction}
\label{sec:intro}

\begin{figure}[t]
    \centering
    \includegraphics[width=0.43\textwidth]{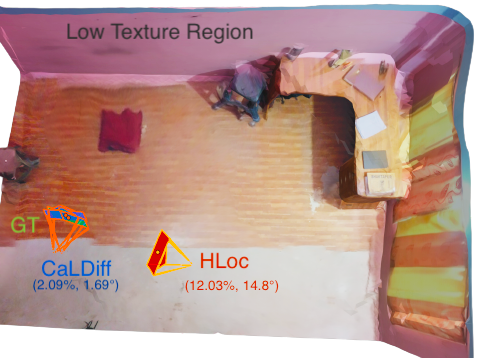}
    \caption{\textbf{Visual Localization using CaLDiff} in a challenging environment. \textcolor{green}{\textbf{Green}} camera represents the ground truth pose, \textcolor{blue}{\textbf{Blue}} represents the pose estimation by CaLDiff algorithm, \textcolor{red}{\textbf{Red}} camera represents pose estimation by hierarchical localization (DISK+lightglue) algorithm. Results from Empty Rooms dataset.}
    \label{fig:teaser}
    \vspace{-5mm}

\end{figure}

Visual localization is the process of determining the position of a camera within a known scene. It is a crucial component for a wide range of applications, including autonomous vehicle navigation, mobile robotics, augmented reality, and Structure-from-Motion (SfM). In the past decades, localization techniques have been intensively investigated for various types of data, such as 3D point clouds~\cite{shi2023lidar}, SfM maps~\cite{sarlin2019coarse}, and posed images~\cite{arandjelovic2016netvlad}. However, solutions tailored for implicit scene representations, like Neural Radiance Fields (NeRF), have received relatively limited attention. 

NeRF~\cite{mildenhall2021nerf} and its variants~\cite{martinbrualla2020nerfw,barron2021mipnerf, muller2022instant, nerfstudio2023}, learn the structure of a scene from pose-annotated image collections, capturing complex 3D structures and high-fidelity surface details. Unlike traditional 3D scene representations, NeRF offers a compact solution, encapsulating the entirety of a scene without auxiliary data such as images, point clouds, or 3D meshes. This standalone capability and the accuracy of NeRF make it a desirable support for applications in various fields. 
Given this context, the relevance of robust visual localization within NeRF-based reconstruction becomes increasingly apparent. While some approaches have been developed to refine a camera localization in a NeRF given a rough estimate of its pose~\cite{lin2021barf,jeong2021self,yen2020inerf}, few techniques have addressed the particular problem of visual localization of a query image in a trained NeRF model without any prior.
For this use case, existing solutions designed for different representations cannot be directly applied without major data pre-processing or additional information. To cope with this limitation, in this paper, we aim to develop a had-doc approach, taking advantage of the specificity of NeRF to perform a robust and accurate 6DoF camera localization. 

In particular, we propose to leverage the image synthesis capability of NeRF to train a diffusion model conditioned upon a query image to be localized in the scene. 
After training such a model using image-pose pairs, the localization is achieved by randomly spreading particles (camera pose candidates) in the scene and iteratively refining them through the diffusion model in order to provide a close estimate of the actual camera pose used to condition the diffusion model. Unlike competing feature-based techniques~\cite{sarlin2019coarse},  this direct localization strategy is particularly effective under challenging conditions such as low-texture environments, making our technique desirable for indoor scenarios (see \figureautorefname~\ref{fig:teaser}). Moreover, such kind of scene-specific localization technique~\cite{kendall2015posenet} is particularly well-suited for NeRF, which is also trained on a single scene. 

While our diffusion itself is able to provide a reliable estimate of the camera pose, it's accuracy remains limited. To address this issue and improve its convergence, we propose to include a final refinement process via a differentiable photometric registration method called iNeRF. 
After the diffusion of each particle is achieved, most of them converge to the query image's location. Finally, given their photometric errors, the top-n particles are used as an initialization to be refined via iNeRF.

Through a large set of experiments, we highlight that this approach not only simplifies the visual localization in a NeRF but also drastically improves the robustness of the pose estimation process by a large margin when tested on challenging indoor scenes where state-of-the-art hierarchical localization techniques typically fail.

This work makes the following contributions:
\begin{itemize}
    \item We propose a novel method to localize images in the neural radiance field, without requiring any initialization, whatsoever. The proposed method uses a pose diffusion model to iteratively search the optimal pose.
    \item Our pose diffusion process is aware of the downstream refinement.
    We enable this capability by integrating the pose inversion steps within the diffusion process.
    \item CaLDiff has a significant advantage over the feature-based hierarchical localization approach, especially for challenging environments like poorly textured scenes.
\end{itemize}

\section{Related Works}

\begin{figure*}[t]
    \centering
    \includegraphics[width=\textwidth]{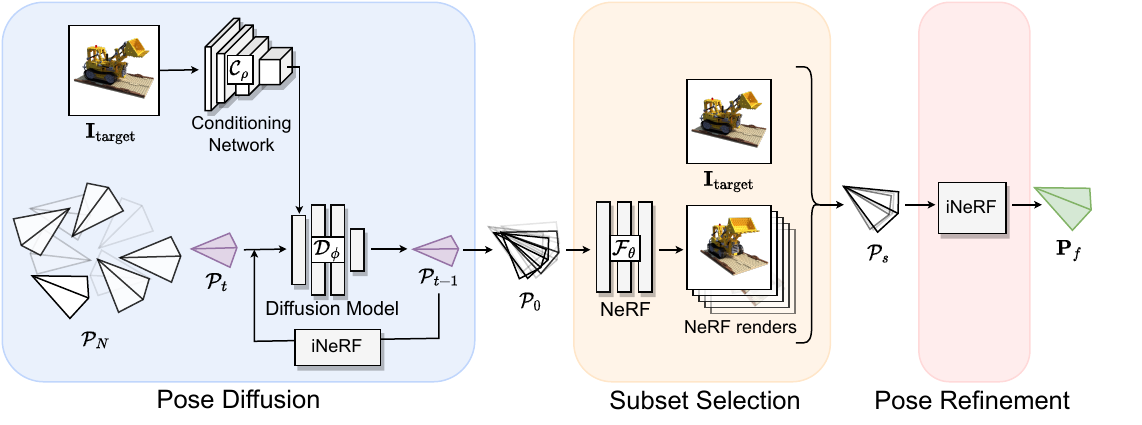}
    \caption{CaLDiff block diagram. \textbf{Pose Diffusion block} takes in initial random poses $\mathcal{P}_N$ and gives rough estimates $\mathcal{P}_0$. Here, $\mathcal{P}_t$ and $\mathcal{P}_{t-1}$ shown by \textcolor{mymagenta}{\textbf{magenta}} camera symbols, represents group of poses at the timestep $t$ and $t-1$ respectively. \textbf{Subset Selection block} outputs $\mathcal{P}_s$ by choosing best poses from $\mathcal{P}_0$. \textbf{Pose Refinement block} futher refines $\mathcal{P}_s$ and gives the best estimte $\mathbf{P}_f$.}
    \label{fig:block_diagram}
\end{figure*}

Visual localization enables the estimation of a camera's orientation and position within a known mapped environment~\cite{piasco2018survey}. The scale and nature of the map utilized for localization can vary significantly depending on the use case. While context-specific maps, such as road maps~\cite{sarlin2023orienternet}, have been investigated, the most general framework involves a prior 3D reconstruction from structure-from-motion techniques~\cite{schonberger2016structure}. In this work, we investigate another medium in which a camera can be localized, namely, a NeRF. This section will introduce the most common localization techniques in explicit 3D representation, followed by emerging strategies to precisely localize a camera from a single query image within an implicit scene reconstruction.
\paragraph{Traditional Visual Localization}
The localization process can operate at varying levels of granularity, providing either an approximate position of the camera or an exact 6DoF pose estimate. The most effective and accurate techniques employ a hierarchical strategy~\cite{sarlin2019coarse}, where the localization is approached in two successive and complementary steps: initially, a coarse localization is achieved using an image retrieval method~\cite{arandjelovic2016netvlad} to narrow the search range. This is followed by a fine pose estimation, which involves a local descriptor-matching strategy~\cite{sarlin2020superglue, detone2018superpoint, tyszkiewicz2020disk}. While these approaches have demonstrated relevance in challenging outdoor environments~\cite{sattler2018benchmarking}, their accuracy remains limited in indoor, textureless environments~\cite{walch17spatiallstms}. In contrast, end-to-end deep learning-based approaches, such as PoseNet~\cite{kendall2015posenet}, have shown better resilience under such conditions. However, these solutions are known to be significantly less accurate than their keypoint-based counterparts.
Moreover, they have the disadvantage of being scene-dependent, meaning that a new network needs to be trained for every new scene. While this could be a limitation in a general setup, it does not pose a significant problem in the particular context of NeRF, as this type of representation is also scene-dependent and requires prior training. Nonetheless, such end-to-end techniques remain rarely used in practice due to their lack of accuracy and explainability. In this paper, we proposed a more advanced approach using diffusion-based pose estimation designed for NeRF that demonstrates better robustness and accuracy than existing techniques under complex scenarios. 
%
%
\paragraph{Pose refinement in NeRF}
Pose estimation solutions have been proposed from the early stage of 3D scene neural representation. Many of them, such as self-calibrating NeRF~\cite{jeong2021self} have been mainly designed to deal with inaccurate pose initialization during NeRF training. These techniques ensure the convergence of the implicit representation model training even when the initial poses lack accuracy. While these methods appear relevant to improve the training of a NeRF they have not originally been developed to address the particular localization problem. One pioneer in this field is iNeRF~\cite{yen2020inerf}, which proposes regressing the camera poses based on an existing NeRF scene representation. It uses gradient descent to minimize the photometric error between pixels rendered from NeRF and the target image to optimize the initial pose estimate. It highlighted that the vicinity of convergence for pose estimation is narrow, akin to direct image alignment methods. 
Building upon this, Barf~\cite{lin2021barf} drastically improves the range of convergence by gradually introducing higher frequency components through a modular positional encoding strategy.
It is noticeable that NeRF-based camera pose refinement has been widely used for real-time SLAM applications such as iMAP~\cite{sucar2021imap} and its successors~\cite{zhu2022nice,zhu2023nicer}.
While relying on similar camera regression strategies for refinement, our proposed method overcomes the limitation of needing an accurate initial camera pose, enabling effective convergence without any prior knowledge of camera localization.
%
\paragraph{NeRF-based visual localization}
With the recent widespread adoption of NeRF for various tasks, the need to localize a camera within such a representation has become increasingly important. However, only a few approaches have been specifically developed for this scenario. Among the existing techniques, we can mention~\cite{maggio2023loc}, which introduces a particle-based localization strategy in a NeRF volume. In this paper, the authors propose localizing a wheeled robot by dispersing 3DoF particles in the robot's environment. While this strategy appears to be effective in such scenarios, it cannot be generalized to more complex situations where the camera to be localized may undergo more complex motions. This major limitation is related to the poor range of convergence of iNeRf used in this technique. To avoid this issue, our strategy relies instead on a diffusion model able to localize a camera in a 6DoF environment effectively. 
Another notable technique is~\cite{liu2023nerfloc}, which proposes to utilize a generalizable and conditional NeRF for 3D scene representation. This approach allows for accurate localization by matching 3D descriptors generated by the NeRF model with 2D image features -- extracted from the query image. This technique includes an appearance adaptation layer to handle domain gaps and improve robustness against varying conditions between the training and testing phases. While this approach demonstrates appealing results, it requires reference posed images with their respective depth map to fine-tune the generalizable NeRF. In contrast, our strategy only necessitates a single pre-train NeRF of any type.
Some other work, such as~\cite{hou2022implicit}, proposes to take advantage of NeRFs to enhance sparse maps with implicit scene representations to traditional improve visual relocalization. This approach, which leverages NeRF, allows for better localization performance in challenging views by removing noisy 3D points and adding back missing details omitted in sparse reconstructions. While such a technique is used to improve conventional 3D maps, our solution has been specifically designed to work on NeRF without additional pre-processing.

\section{Method}
\label{sec:method}

\subsection{Problem Formulation}
Given a monocular RGB image $\mathbf{I}$, our method provides 6DoF pose $\mathbf{P}=[\mathbf{x},\mathbf{q}]$ of the camera, given by 3D translation vector $\mathbf{x}$ and orientation represent by 4D quaternion $\mathbf{q}$.
We formulate the problem of pose estimation as a two-step process. First, find rough pose estimates using pose diffusion, then refine it using iNeRF optimization.

\subsection{Method Overview}
The frame of proposed CaLDiff method for camera localization in NeRF is illustrated in Figure~\ref{fig:block_diagram}. As shown, we use three different steps to obtain the desired camera pose parameters. In the first step, the target image is localized coarsely using the proposed pose diffusion method. These coarse poses are then further processed by a local refinement using the iNeRF optimization~\cite{yen2020inerf}. Due to the particle-like nature of the diffusion process, we obtain many coarse pose proposals, a subset of them is then selected for the further refinement. Further details of each of these three steps are provided in the Subsections~\ref{sub:posDiff},~\ref{sub:subset}, and~\ref{sub:iNerfRef}, respectively. We provide the necessary developments before presenting there subsections below.   

\subsection{Pose Initialization}
To facilitate the stochastic behaviour of the diffusion process, we sample $P$ random poses $\mathcal{P}_N=\{\mathbf{P}_N^i\}_{i=0}^P$ from uniform distribution in the normalized coordinates. We sample translation vector $\mathbf{x} \sim U(-0.5,0.5)^3$ from uniform distribution in $\mathbb{R}^3$. We sample a random point in unit sphere for quaternion vector $\mathbf{q}$. We take $u,v,w \sim U(0,1)$ and sample $\mathbf{q}$ as,
\begin{equation}
\begin{aligned}
    \mathbf{q} &= (\sqrt{1-u} \sin(2\pi v), \sqrt{1-u} \cos(2\pi v), \\
    &\quad \sqrt{u} \sin(2\pi w), \sqrt{u} \cos(2\pi w)).
\end{aligned}
\end{equation}
Each of these pose will independently diffuse during pose diffusion, over the course of multiple diffusion steps.  

\subsection{Conditioning Features}
To meaningfully diffuse the randomly selected poses, towards the pose of the target image (i.e. the image to be localized), we condition diffusion model using feature vector $\mathbf{f}$. This feature vector is generated by the conditioning network $\mathcal{C}_\rho$ using the target image $\mathbf{I}_\text{target}$ as,
$$\mathcal{C}_\rho(\mathbf{I}_\text{target})\rightarrow\mathbf{f}.$$
In our experiment, we use image features obtained form the ResNet~\cite{he2015deep} train on the ImageNet~\cite{DenDon09Imagenet} dataset.

\subsection{Pose Diffusion}
\label{sub:posDiff}
During the pose diffusion step, each pose $\mathbf{P}^i$ undergoes $N$ reverse diffusion steps via the diffusion model $\mathcal{D}_\phi$, conditioned on the feature vector $\mathbf{f}$ as, $$\mathcal{D}_\phi(\mathcal{P}_t,\mathbf{f},t)\rightarrow \mathcal{P}_{t-1}. $$
We apply sinusoidal positional embedding $\gamma$ on each element of vector $\mathbf{P}^i$ and the time timestamp $t$. Input to the MLP of $\mathcal{D}_\phi$ is the concatenated $\mathbf{f}$, $\gamma(\mathbf{\mathcal{P}}_t)$, and $\gamma(t)$. We do quaternion normalization on the output of the network to get $\mathcal{P}_{t-1}$, which is then fed back to the same network for iterative refinement until $t=0$. Quaternion normalization is necessary because the diffusion process doesn't work on $\mathfrak{se}(3)$ manifold and the quaternion vector needs to be normalized to represent true rotation.

During training, we corrupt a target pose $\mathbf{P}_\text{target}$ by adding noise for a given timestamp $t$, guided by the forward diffusion equation \ref{eq:fd}. 
\begin{equation}
    \begin{aligned}
        \mathbf{P}_t &= \sqrt {\bar \alpha_t} \mathbf{P}_0 + \sqrt{1-\bar \alpha_t} \epsilon \\
    \end{aligned}
    \label{eq:fd}
\end{equation}
$\beta_t\in(0,1)$ is the amount of noise added at step $t$ w.r.t step $t-1$. $\alpha_t = 1-\beta_t$ is the amount of information preserved. $\bar \alpha_t = \prod_{i=1}^{t} \alpha_i$ is the amount of image information preserved at the step $t$ w.r.t step 0. And, $\epsilon \in (0,1)$ is random noise. We use the reverse diffusion equation \ref{eq:rd} to get $\mathbf{P}_{t-1}.$
\begin{equation}
    \begin{aligned}
        \mathbf{P}_{t-1} = \frac{1}{\sqrt{\alpha_t}} \left(\mathbf{P}_t - \frac{1-\alpha_t}{\sqrt{1-\bar \alpha_t}} \psi(\mathbf{P}_t,t) \right) + (1-\alpha_t) \epsilon
    \end{aligned}
    \label{eq:rd}
\end{equation}
$\psi$ represents the diffusion model MLP which predicts the noise added to the sample $\mathbf{P}_0$ for a given time step $t$. We compare $\mathbf{P}_{t-1}$ with the target pose $\mathbf{P}_\text{target}$ to get the error used to train $\psi$. We use balanced mean square error (MSE) loss represented by the equation \ref{eq:ls} which balances translational and rotational MSE error with balance factor $\lambda$.
\begin{equation}
        \text{balanced loss}=\text{mse}(\mathbf{x}_{t-1},\mathbf{x}_\text{target}) + \lambda . \text{mse}(\mathbf{q}_{t-1},\mathbf{q}_\text{target})
    \label{eq:ls}
\end{equation}

Here, the timestep $t\in \{ N-1, N-2, ..., 1, 0 \} $. Final estimates $\mathcal{P}_0$ are the rough estimates of this step. 

\subsection{Subset Selection}
\label{sub:subset}
For the computational reasons, we select $B$ best poses ${\mathcal{P}_S}={\{\mathbf{P}_S^i\}}_{i=1}^{B}$ from $\mathcal{P}_0$ based on least photometric error with respect to target image $\mathbf{I}_{\text{target}}$, and apply iNeRF optimization on them. Photometric loss function $\mathcal{L}$ calculates pixel-wise mean square error $\mathbf{e}$ between $\mathbf{I}_{\text{target}}$ and NeRF renders with the poses $\mathcal{P}_S$ as represented by Equation \eqref{eq:pl}.
\begin{equation}
        \mathbf{e}=\mathcal{L}( \mathcal{\mathcal{F}_\theta}(\mathcal{P}_S))  ,  \mathbf{I}_{\text{target}} ).
    \label{eq:pl}
\end{equation}
The selected poses are the ones which result into the minimum erros $\mathbf{e}$. Please recall that we are eventually interested to find poses that best minimizes the photometric loss. 

\subsection{Pose Refinement}
\label{sub:iNerfRef}
During the NeRF training, we train the model $\mathcal{F}_\theta$ for a scene on image-pose pairs $\mathcal{T}=\{ (\mathbf{I}^j,\mathbf{P}^j)_{j=1}^{T} \}$ having $T$ training samples. $\mathcal{F}_\theta$ generalizes over the scene and learns to render views from unseen poses. Given a pre-trained NeRF model, a target image, and the coarse pose initialization obtained after the pose diffusion, now we are interested to refine the pose further. Following~\cite{yen2020inerf}, we refine each pose $\mathbf{P}_S$ by minimizing the photometric loss $\mathcal{L}$ in Equation~\eqref{eq:inerf}, to get best estimate $\hat{\mathbf{P}}_S$.
\begin{equation}
    \hat{\mathbf{P}}_S=\arg\min_{\mathbf{P}_S} \mathcal{L}(\mathcal{F}_\theta(\mathbf{P}_S), \mathbf{I}_{\text{target}}).
    \label{eq:inerf}
\end{equation}
For simplicity, among estimates $\{\hat{\mathbf{P}}_S^i\}_{i=1}^{B}$ corresponding to the previously selected set $B$, we select the one with least photometric error $\mathcal{L}(\mathcal{F}_\theta(\hat{\mathbf{P}}_S^i), \mathbf{I}_\text{target})$. Let the finally selected refined pose be $\mathbf{P}_f$. Our results reported throughout this paper correspond to $\mathbf{P}_f$, unless mentioned otherwise.

\subsection{Pose Diffusion with iNeRF steps}

We integrate the inverse NeRF step in Equation~\eqref{eq:inerf} within reverse diffusion step in Equation~\eqref{eq:rd}. We obtain $\bar{\mathcal{P}}_{t-1}$ from $\mathcal{P}_t$ using reverse diffusion step of $\mathcal{D}_\phi$. During the integrated iNeRF diffusion, we do not directly feedback  $\bar{\mathcal{P}}_{t-1}$ to the network $\mathcal{D}_\phi$. Instead, we apply M steps of iNeRf optimization steps, as presented in the Subsection~\ref{sub:iNerfRef}. We apply iNeRF steps on each poses $\bar{\mathbf{P}}_{t-1}\in\bar{\mathcal{P}}_{t-1}$ to obtain the refined poses $\mathcal{P}_{t-1}$.  The pose evolution during this process is shown below,
\begin{tikzpicture}
\node at (0,0) {$\mathcal{P}_t$};
\draw[->, thick] (0.3,0) -- (1.8,0);
\node at (2.3,0) {$\bar{\mathcal{P}}_{t-1}$};
\node at (1.0,0.3) {\scriptsize $\mathcal{D}_\phi$};
\draw[->, thick] (2.8,0) -- (4.0,0);
\node at (4.5,0) {${\mathcal{P}}_{t-1}$};
\node at (3.4,0.3) {\scriptsize iNerf};
\node at (5.4,0.3) {\scriptsize $\mathcal{D}_\phi$};
\draw[->, thick] (5,0) -- (6.5,0.0);
\node at (7.0,0.0) {$\ldots$};
\end{tikzpicture}
\\
The choice of such pose evolution is made to make the diffusion process aware of the downstream refinement process. 
Another reason also involves it simplicity to integrate. Please, refer to Figure~\ref{fig:block_diagram} (on the left) for the schematics of the iNeRF integration.

\subsection{Network Architecture}

 Diffusion Model consists of linear layers with 5 hidden layers of size 1024 and output layer of size 7. The 7 dimensional input and time step input of diffusion model is encoded using sinusoidal embeddings before feding into the network. We use ResNet18~\cite{he2015deep} pre-trained on ImageNet~\cite{DenDon09Imagenet} dataset as the Conditioning Network.

We use Nerfacto model provided by Nerfstudio~\cite{nerfstudio2023} for creating NeRF models. It consists of a multi-resolution hash grid with 16 levels, resolution varying from 16 to 2048, paired with a  small fused-MLP with 2 hidden layers of size 64 to represent the scene. Standard NeRF encoding is used for positional encoding while directions are encoded using spherical harmonics. A combination of piece-wise and proposal sampler is used to sample 96 samples per ray. 

\subsection{The CaLDiff Algorithm}

We provide the summary of the proposed CaLDiff localization method in Algorithm~\ref{alg:algorithmTri}.
As can be seen, our method uses the conditioning network $\mathcal{C}_\rho$ to extract feature $\mathbf{f}$. This feature is then used in the diffusion network $\mathcal{D}_\phi$ to diffuse the poses $\mathcal{P}_t$ in step 3. 
This process provides us the coarse poses $\mathcal{P}_0$, from which a subset $\mathcal{P}_s$ of best poses are selected. The final pose $\mathsf{P}_f$ is then obtained after refining $\mathcal{P}_s$, followed by the best pose selection. 

\begin{algorithm}
{
\caption{ $\mathbf{P}_f$ = CaLDiff\_localization ($\mathbf{I}_\text{target}$)}
\begin{algorithmic}
\label{alg:algorithmTri}
 \STATE 1. Initialize poses $\mathcal{P}_N=\{\mathbf{P}_T^i\}_{i=0}^P$.
 \STATE 2. Extract feature $\mathbf{f}=\mathcal{C}_\rho(\mathbf{I}_\text{target})$.
 \STATE 3. Diffuse $\mathcal{P}_N$ to $\mathcal{P}_0\!=\!\{\mathbf{P}_0^i\}$ using $\mathcal{D}_\phi(\mathcal{P}_t,\mathbf{f},t)\!\rightarrow \!\mathcal{P}_{t-1}$.
 \STATE 4. Select the good-to-refine subset $\mathcal{P}_S\subset \mathcal{P}_0$.
 \STATE 5. Refine $\mathcal{P}_S$ using iNeRF to $ \mathcal{P}_f$ and select the best $\mathbf{P}_f$.
 \STATE 6. Return the final refined pose $\mathbf{P}_f$.
\end{algorithmic}
}
\end{algorithm}

\section{Experiments}
\label{sec:exp}
In this section, we propose a large set of experiments on three different datasets to highlight the advantages and limitations of our localization technique. To better analyze the relevance of the different modules composing CaLDiff, we also conducted an in-depth ablation study.

\subsection{Datasets}
For our experiments, we relied on three datasets: Synthetic dataset~\cite{mildenhall2021nerf}, 7scenes~\cite{glocker2013real}, and our Custom dataset 'Empty rooms'. 
For our synthetic experimentations, we used three scenes of the dataset~\cite{mildenhall2021nerf}: Cars, Chair, and Lego, with its multiple subsets for training and testing. 
This synthetic dataset is well suited to perform ablation analysis. Moreover, it comes with its own challenges, such as full $360^{\circ}$ scenes, repetitive patterns, textureless environments, and poorly distributed visual data.
In order to demonstrate the capabilities of CaLDiff under real scenarios and to compare it with existing state-of-the-art techniques, we opted for 7scenes, which is a very commonly used dataset for visual localization in indoor environments. This dataset is composed of seven different indoor scenes with thousands of images per scene, along with their ground truth poses. We select 200 random images for training and 50 for testing.
Finally, to highlight the robustness of CaLDiff under challenging environments, we prepare our own dataset 'Empty Rooms'. This dataset is composed of seven sets of 130 image-depth pairs captured in empty rooms and staircase scenarios. Note that most images captured under this environment contain little to no textures. Therefore, we acquired RGB images along with their LIDAR depth information using an iPad 11 Pro. These depth maps are essential as the image alone is insufficient for reconstructing the scene. Given this information, we used Polycam~\cite{polycam2022} to generate the ground truth poses used to train NeRFs.


\subsection{Model Training and Evaluation Protocols}

\paragraph{Train NeRF model.} 
For the synthetic scenes, we use the tiny NeRF~\cite{mildenhall2021nerf}  model, which is well suited for these minimalist scenarios. Regarding the real scenes, we opted for the nerfacto~\cite{nerfstudio2023} model, which is more appropriate for large scenes. Having different network architectures for evaluation also demonstrates the flexibility of our technique, which can be applied to most NeRF models.
We conducted training for the models using 20k iterations for Tiny NeRF and 10k iterations for NerfActo. During each Tiny NeRF iteration, a single random data point was selected from the training set. In contrast, for NerfActo, each iteration involved performing volumetric rendering on a subset of 4,096 pixels, randomly chosen from the entire dataset.


\paragraph{Train Diffusion model.} We train the Diffusion model $\mathcal{D}_\phi$ along with the conditioning network $\mathcal{C}_\rho$. We use Adam with a learning rate of $0.0002$. We use pose error balance factor $\lambda=2$. For each scene, we train the model for 20 epochs with 5000 iterations per epoch. On each iteration, one pose is randomly selected for the available training poses.

\paragraph{Testing protocol.} We sample 100 random poses from a uniform distribution and run the reverse diffusion process on each pose for $N=50$ timesteps conditioned on the target image. We use linear beta scheduling, $$\beta_t=\beta_0 + t*\frac{\beta_N-\beta_0}{N}.$$ We take $\beta_0=0.0001$ and $\beta_N=0.02$. This generates rough pose estimates. We do NeRF renders for all the estimated poses, and calculate the photometric error w.r.t. the ground truth image. We select the $B=5$ best poses with the least photometric error and apply iNeRF optimization on them.

For iNeRF optimization, the mean square photometric error is backpropagated via the NeRF model to optimize the pose. We use Adam optimizer with a learning rate $0.007$. We do optimization for 300 steps. To reduce the computational cost, we render only a subset of pixels of the entire image. We detect $100$ SIFT keypoints on the target image with a very low detection threshold, to get the points even at highly textureless surfaces. We dilate keypoints locations by 5 pixels and get the subset of pixels. We do NeRF rendering only at these pixel locations and compute photometric loss with the subset pixels of the target image $\mathbf{I}_\text{target}$.

We compare our results with the hierarchical localization technique using DISK~\cite{tyszkiewicz2020disk} feature descriptor and LightGlue~\cite{lindenberger2023lightglue} matcher. For this purpose, we create a 3D map using ground truth poses of train split and use the localization pipeline provided by Hloc~\cite{sarlin2019coarse} to localize the test images in the given 3D map. Also, we replaced the Pose Diffusion step on our method with Monte Carlo-based sampling technique~\cite{lin2023parallel} as one of the baselines.

We report the success rate of localization on 4 thresholds: (0.01,2°), (0.025,5°), (0.05,10°), (0.1,20°). Here, the first element is translation threshold for normalized coordinates and the second is rotational threshold in degrees.

\subsection{Discussion}
The results obtained on the synthetic scenes are available in Table~\ref{tab:synthetic_exp}. They clearly underline the robustness of our technique in such scenarios regardless of the number of images used. This performance gap between the feature-based technique Hloc and ours is due to multiple factors. Indeed, the synthetic images, contain very little textures, many repetitive patterns, and the visual content is non-homogeneously distributed. These combined elements offer very unfavorable conditions for feature-based techniques while photometric approaches such as ours remain effective. 
These results are corroborated by our tests on real-world images which we provide in Table~\ref{tab:realworld_exp}. In this table, we compare CalDiff with three other baselines: Hloc~\cite{sarlin2019coarse}, PoseNet~\cite{kendall2015posenet} and MonteCarlo~\cite{maggio2023loc}.
Unsurprisingly, Hloc shows satisfactory performance on the 7scenes dataset, except in texture-poor scenes such as the stairs. CaLDiff, in contrast, is unaffected by textureless scenes, exhibiting robust performance across the dataset. This resilience can be explained by the combined use of direct pose regression combined with a photometric registration technique. However, PoseNet, despite following a similar pipeline, performs poorly in 7scenes due to its single-stage pose estimation. We can notice that the multi-steps and multi-hypothesis approach of CaLdiff results in a much more accurate and robust pose estimation. 
Finally, the Monte Carlo method, while utilizing multiple pose hypotheses, consistently fails across the dataset. This limitation can be linked to its original design, which was intended for 3DoF robot localization. The higher degree of freedom and the limited convergence range of iNeRF likely result in the Monte Carlo particles failing to approximate the expected position closely enough for effective convergence.
In the more challenging Empty rooms dataset, both Hloc and PoseNet+iNeRF exhibit a marked decrease in performance compared to the previous datasets. Notably, PoseNet+iNeRF encounters complete failures in several scenes. CaLDiff, while also experiencing a reduced success rate relative to its performance on the 7scenes dataset, demonstrates a comparatively better adaptability, maintaining a degree of effectiveness despite the increased difficulty of the dataset.

\begin{table}[]
\resizebox{0.47\textwidth}{!}{%

\begin{tabular}{llll}

\hline

\textbf{Scene} & 
\textbf{Split} & 
\begin{tabular}[c]{@{}l@{}}\textbf{Hloc} $\uparrow$\\ \textbf{(DISK+lightglue)}\end{tabular} & 
\begin{tabular}[c]{@{}l@{}}\textbf{CaLDiff} $\uparrow$ \\ \textbf{(Ours)}\end{tabular} \\ 

\hline

cars  & 50-25 & 3, 9, 13, 15& 32, 57, 63, 65\\
      & 25-25 & 2, 4, 11, 12& 8, 36, 44, 56\\
      & 15-25 & 0, 0, 0, 1& 4, 8, 20, 28\\ \hline
chair & 50-25 & 4, 34, 42, 60                                                   & 52, 72, 72, 84\\
      & 25-25 & 4, 13, 29, 37& 34, 48, 68, 80\\
      & 15-25 & 2, 10,15,20                                                     & 4, 16, 20, 28\\ \hline
lego  & 50-25 & 8, 34, 50, 56                                                   & 80, 84, 84, 88\\
      & 25-25 & 6, 26, 36,48                                                    & 42, 67, 73, 81\\
      & 15-25 & 0, 2, 10, 24                                                    & 5, 12, 44, 63\\ \hline
\end{tabular}}

\caption{\textbf{Results on synthetic dataset.} We report the success rate of localization on 4 thresholds: (0.01,2°), (0.025,5°), (0.05,10°), (0.1,20°). Here, first element is translation threshold for normalized coordinates and second is rotational threshold in degrees. For each scene, we take 3 different train-test splits having 50-25, 25-25, and 15-25 images. We take 5 variants of each type of train-test splits and report the average result among them.}
\label{tab:synthetic_exp}
\vspace{-5mm}

\end{table}

\begin{table*}[]
\centering
\begin{tabular}{llllll}
\hline
\textbf{Dataset} & \textbf{Scene} & \textbf{Hloc (Disk+lightglue)} $\uparrow$ & \textbf{PoseNet+iNeRF} $\uparrow$  & \textbf{Monte Carlo + iNeRF} $\uparrow$ & \textbf{CaLDiff (Ours)} $\uparrow$ \\ \hline
&chess              & 10, 60, 80, 86                                 & 8, 30, 30, 44                    &0, 0, 0, 0&    \textbf{40, 68, 88, 96}\\
&fire               & 14, 34, 62, 76,                                & 17, 18, 19, 2                    &0, 0, 0, 0&    \textbf{64, 84, 84, 84}\\
&heads              & 16, 57, 74, 86                                 & 16, 20, 22, 28                   &0, 0, 0, 0&    \textbf{73, 86, 86, 86}\\
7scenes &office     & 14, 40, 44, 44                                 & 11, 18, 18, 9                    &0, 0, 0, 0&    \textbf{52, 96, 96, 96}\\
&pumpkin            & 0, 20, 48, 70                                  & 6, 17, 26, 31                    &0, 0, 0, 0&    \textbf{14, 54, 77, 77}\\
&redkitchen         & 12, 52, 68, 70                                 & 4, 9, 10, 15                     &0, 0, 0, 0&    \textbf{25, 75, 79, 86}\\
&stairs             & 0, 6, 8, 32                                    & 1, 2, 2, 5                       &0, 0, 0, 0&    \textbf{58, 58, 58, 75}\\ 
\hline
&room1              & 10, 20, 20, 20                                 &0, 0, 0, 0                        &0, 0, 0, 0&    \textbf{23, 30, 30, 33}          \\
&room2              & \textbf{3, 3,} 3, 10                                   &0, 0, 0, 0                         &0, 0, 0, 0&    \textbf{3, 3, 30, 50}           \\
&room3              & 0, 0, 0, 3                                  &0, 0, 0, 0                            &0, 0, 0, 0&    \textbf{7, 17, 27, 37}          \\
Empty&room4         & 4, 14, 14, 22                                  &0, 8, 9, 11                       &0, 0, 0, 0&    \textbf{22, 47, 70, 87}          \\
Rooms&room5         & 6, 10, 12, 12                                  &1, 1, 1, 5                        &0, 0, 0, 0&    \textbf{27, 58, 72, 87}          \\
&room6              & 5, 17, 22, 22                                  &2, 6, 7, 8                        &0, 0, 0, 0&    \textbf{35, 66, 80, 85 }         \\ 
&stairs1            & 3, 6, 6, 100                                   &1, 1, 1, 2                        &0, 0, 0, 0&    \textbf{30, 45, 51, 66}          \\
\hline
\end{tabular}
\vspace{-2mm}
\caption{\textbf{Evaluation on Real dataset.} We compare CaLDiff with 3 benchmarks hloc~\cite{sarlin2019coarse}, PoseNet~\cite{kendall2015posenet}+iNeRF~\cite{yen2020inerf} and Monte Carlo~\cite{lin2023parallel}+iNeRF based approaches. The best method in the row is in \textbf{Bold}. CaLDiff works better on all the cases. }
\label{tab:realworld_exp}
\end{table*}

\subsection{Qualitative results}

\begin{figure}[t]
    \centering
    \includegraphics[width=0.11\textwidth]{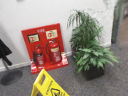}
    \includegraphics[width=0.11\textwidth]{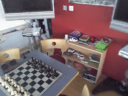}
    \includegraphics[width=0.11\textwidth]{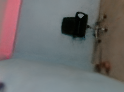}
    \includegraphics[width=0.11\textwidth]{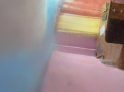}
    \includegraphics[width=0.11\textwidth]{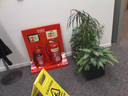}
    \includegraphics[width=0.11\textwidth]{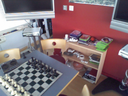}
    \includegraphics[width=0.11\textwidth]{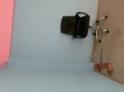}
    \includegraphics[width=0.11\textwidth]{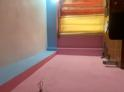}
    \caption{Image rendered using the pose from CaLDiff (top), and the corresponding ground-truth (bottom). Images from 7scenes (left two) and Empty Rooms (right two), results in Table~\ref{tab:realworld_exp}.}
    \vspace{-5mm}
    \label{fig:vis}
\end{figure}

\begin{figure*}[t]
    \centering
    \includegraphics[width=\textwidth]{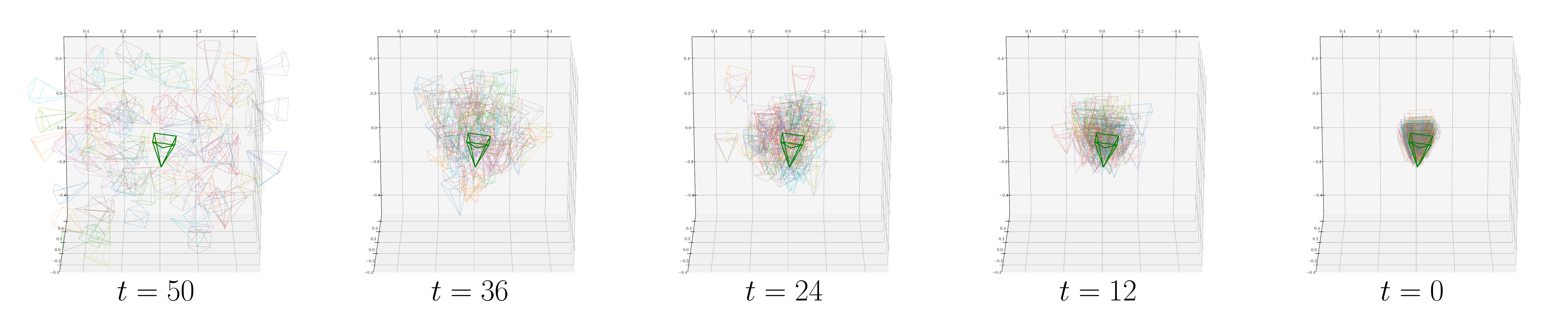}
    \caption{\textbf{Pose diffusion step.} Bold \textcolor{mygreen}{\textbf{green}} camera represents the ground truth pose. We run reverse diffusion for $N=50$ steps. Initial particles at $t=50$ are completely random and slowly converge toward the ground truth pose as $t$ decreases. Finally, at $t=0$ the particles are tightly clustered around the ground truth pose.}
    \label{fig:pose_diffusion}
\end{figure*}

We present the results for the reverse diffusion step for a single pose. Figure~\ref{fig:pose_diffusion} shows how the initial random particles at $t=50$ move towards the ground truth pose as the timestamp decreases to $0$. The bold green camera at center represents the ground truth pose and the faded one represent the diffusion particle at the given timestamp $t$. In Figure~\ref{fig:vis}, we provide some example rendered images. 

\subsection{Ablation Study}
In order to evaluate the individual contribution of each module in CaLDiff, we propose an ablation study where the technique is run with diffusion only and with the final refinement module. This test has been conducted on the synthetic dataset for three different scenes and with varying numbers of images to demonstrate the ability of our localization strategy to perform well even under sparse training image datasets. The obtained results are visible in Table~\ref{tab:ablation_synthetic}. We can notice that the diffusion alone is able to provide a rough localization estimation for most scenarios, for instance, with 56\% of the images being approximately localized for the chair sequence. This initial pose estimation is sufficient to reach the vicinity of convergence of our iNeRF refinement stage such that most scenes are accurately localized. Regarding the resilience of our strategy to the number and distribution of the training data, we can notice a clear decline in accuracy when the training set is being decimated. Nonetheless, despite such challenging conditions, it remains able to localize a large number of testing images. Similar conclusions are observable in real data, as illustrated in Table \ref{tab:ablation_real}.

\begin{table}[]
\begin{tabular}{llll}
\hline
\textbf{Scene} & \textbf{Split} & \begin{tabular}[c]{@{}l@{}}\textbf{CaLDiff (Pose} \\ \textbf{Diffusion only)}\end{tabular} &\begin{tabular}[c]{@{}l@{}}\textbf{CaLDiff (without} \\ \textbf{integrated iNeRF)}\end{tabular}\\ \hline
cars  & 50-25 & 0, 1, 7, 28                                                        &29, 51, 61, 65                                    \\
      & 25-25 & 0, 0, 6, 20                                                              &8, 24, 36, 46                                     \\
      & 15-25 & 0, 0, 0, 3                                                             &4, 11, 15, 23                                    \\ \hline
chair & 50-25 & 0, 4, 18, 56                                                             &48, 76, 82, 90                                 \\
      & 25-25 & 0, 0, 11, 41                                                         &19, 30, 52, 64                                \\
      & 15-25 & 0, 0, 2, 10                                                              &4, 5, 13, 29                                    \\ \hline
lego  & 50-25 & 0, 4, 24, 60                                                             &76, 85, 87, 88                                    \\
      & 25-25 & 0, 1, 17, 61                                                         &44, 66, 74, 82                                   \\
      & 15-25 & 0, 0, 8, 26                                                              &6, 14, 32, 54                 \\ \hline
\end{tabular}
\caption{\textbf{Ablation Study on Synthetic Dataset.} We run two ablation studies on synthetic scenes. The first one discards the post-refinement step and runs Pose Diffusion only. The second one runs the exact process as in Figure ~\ref{fig:block_diagram}, but omits the integrated iNeRF block within the Pose Diffusion Block.}
\label{tab:ablation_synthetic}
\end{table}

\begin{table}[]
\begin{tabular}{ll|ll}
\hline
\multicolumn{2}{l|}{\textbf{Empty Rooms}} & \multicolumn{2}{l}{\textbf{7scenes}} \\ \hline
\textbf{Scene}        & \textbf{Success Rate}      & \textbf{Scene}       & \textbf{Success Rate}  \\ \hline
room1        & 0, 0, 7, 17       & chess       & 0, 8, 36, 84               \\
room2        & 0, 0, 0, 7        & fire        & 0, 8, 24, 64
              \\
room3        & 0, 0, 3, 27       & heads       & 0, 9, 13, 40
              \\
room4        & 0, 0, 4, 48       & office      &  0, 8, 24, 76             \\
room5        & 0, 0, 10, 48      & pumpkin     &  0, 0, 9, 42              \\
room6        & 0, 4, 19, 53      & redkitchen  &   0, 7, 21, 68            \\
stairs1      & 0, 0, 4, 48       & stairs      & 0, 8, 8, 33  \\ \hline
\end{tabular}
\caption{\textbf{Ablation Study on Real data.} We removed the pose refinement process and report the results form the pose diffusion steps only. The accurcy is highly impared both for 7scenes and Empty rooms dataset.}
\label{tab:ablation_real}
\end{table}

\subsection{Behaviour Study}


\begin{figure}[t]
    \centering
    \begin{subfigure}[b]{0.23\textwidth}
        \centering
        \includegraphics[width=\textwidth]{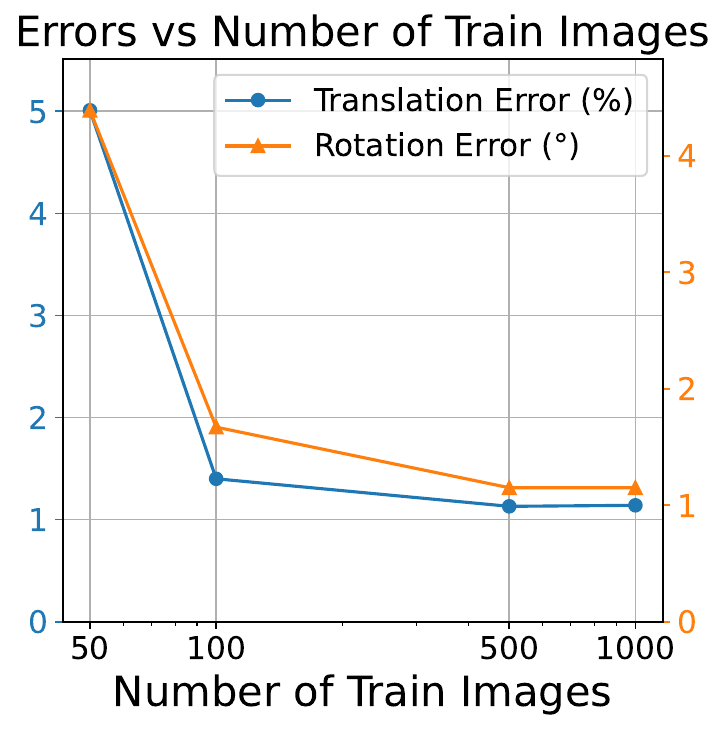}
        \caption{}
        \label{fig:sub1}
    \end{subfigure}
    \hfill
    \begin{subfigure}[b]{0.23\textwidth}
        \centering
        \includegraphics[width=\textwidth]{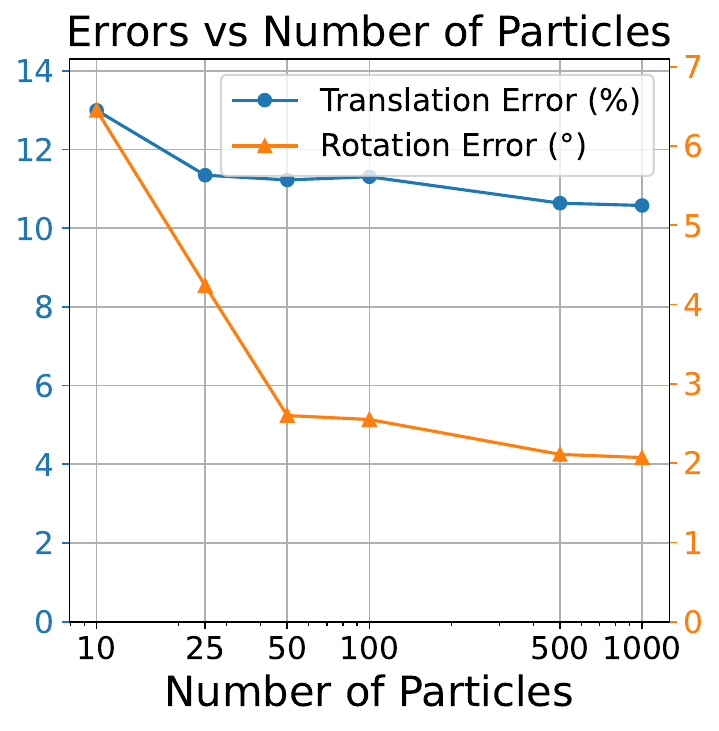}
        \caption{}
        \label{fig:sub2}
    \end{subfigure}
    \begin{subfigure}[b]{0.23\textwidth}
        \centering
        \includegraphics[width=\textwidth]{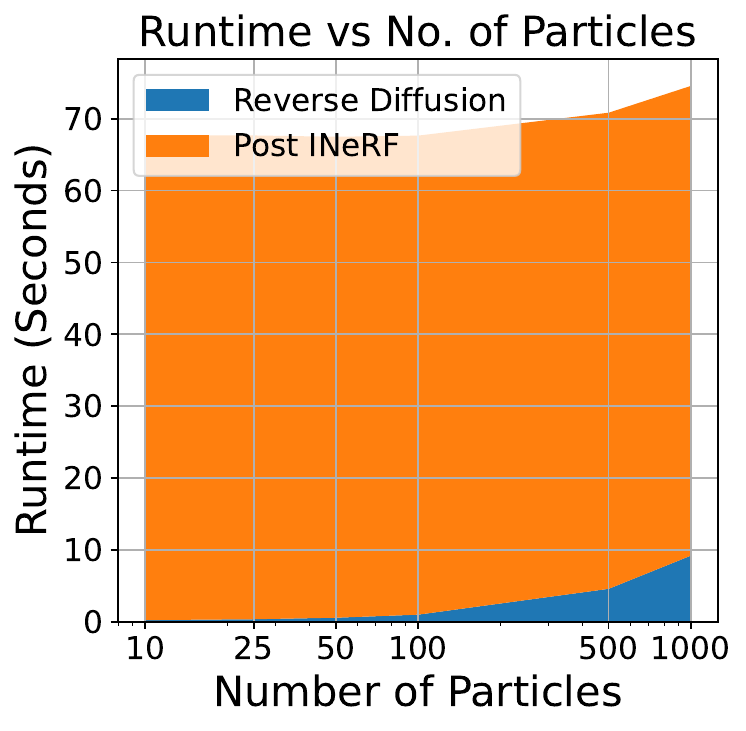}
        \caption{}
        \label{fig:sub3}
    \end{subfigure}
    \hfill
    \begin{subfigure}[b]{0.23\textwidth}
        \centering
        \includegraphics[width=\textwidth]{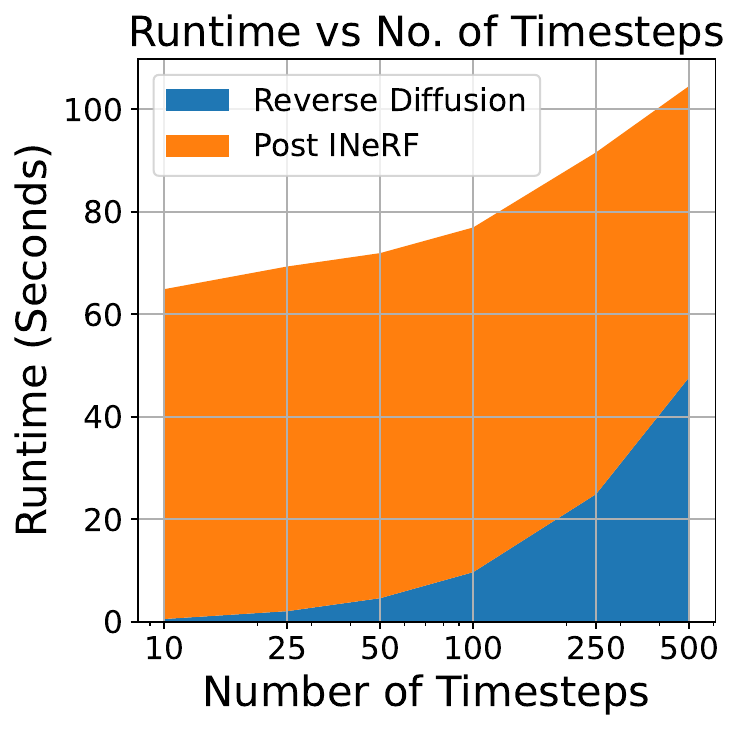}
        \caption{}
        \label{fig:sub4}
    \end{subfigure}
    \vspace{-2mm}
    \caption{\textbf{Behaviour Study.} Shows the effect of number of diffusion particles $P$, number of train images $T$ and number of reverse diffusion steps $N$ on localization error and execution time.}
    \label{fig:behaviour}
    \vspace{-5mm}
\end{figure}

We study how the localization error and execution time varies on changing the number of diffusion particles $P$, number of train images $T$ and number of reverse diffusion steps $N$. We conduct all the experiments on the Chess scene from the 7-scenes dataset with 200-50 train-test split. We run these experiments on an nvidia RTX 4080 GPU. An increase in the number of train images has helped a lot to decrease the localization error as seen in Figure~\ref{fig:sub1}. More specifically, when the number of train images increases from 50 to 100. A similar trend can be observed in Table~\ref{tab:synthetic_exp} when the number of train images increases from 15 to 25.

Increasing the number of particles shows a marked influence on the localization accuracy as seen in Figure~\ref{fig:sub2}, but the returns are diminishing. Figure~\ref{fig:sub3} The runtime is not greatly influenced by increasing the number of particles as the majority of time is taken by post-iNeRF step, which remains constant for any number of initial particles. We plot in Figure~\ref{fig:sub4}, the median localization error and runtimes for a few different numbers of timesteps in the diffusion process. The time taken for localization also increases due to the reverse diffusion process taking longer time.

\section{Conclusion}
\label{sec:conclusion}
In this work, we studied the problem of localizing cameras in the implicitly represented 3D scenes.
The implicit nature of the representation makes the use of existing methods difficult. 
Therefore, we proposed a novel camera localization method that benefits from the recent success in the diffusion based methods. Our diffusion model diffuses randomly initialized camera poses conditioned upon the features of the target image to be localized. The conditioned diffusion model learns the pose gradient field to guide the random poses towards the target's location, in an iterative process. We also showed that the integration of explicit gradient steps leads to better results. The experiments conducted on both synthetic and real datasets provide superior results when compared to the state-of-the-art methods and the established baselines, especially when the challenging cases of the poor texture images are considered.   

\paragraph{Limitations:} One limitation of our method is the limited speed for some real-time applications. This requirements maybe addressed by application/hardware-specific code optimisation and the system configurations. Some aspects of speeding up our method remains as our future works.

\clearpage
{\small
\bibliographystyle{unsrt} 
\bibliography{egbib}

\begin{thebibliography}{10}

\bibitem{shi2023lidar}
Pengcheng Shi, Yongjun Zhang, and Jiayuan Li.
\newblock Lidar-based place recognition for autonomous driving: A survey.
\newblock {\em arXiv preprint arXiv:2306.10561}, 2023.

\bibitem{sarlin2019coarse}
Paul-Edouard Sarlin, Cesar Cadena, Roland Siegwart, and Marcin Dymczyk.
\newblock From coarse to fine: Robust hierarchical localization at large scale.
\newblock In {\em CVPR}, 2019.

\bibitem{arandjelovic2016netvlad}
Relja Arandjelovic, Petr Gronat, Akihiko Torii, Tomas Pajdla, and Josef Sivic.
\newblock Netvlad: Cnn architecture for weakly supervised place recognition.
\newblock In {\em Proceedings of the IEEE conference on computer vision and pattern recognition}, pages 5297--5307, 2016.

\bibitem{mildenhall2021nerf}
Ben Mildenhall, Pratul~P Srinivasan, Matthew Tancik, Jonathan~T Barron, Ravi Ramamoorthi, and Ren Ng.
\newblock Nerf: Representing scenes as neural radiance fields for view synthesis.
\newblock {\em Communications of the ACM}, 65(1):99--106, 2021.

\bibitem{martinbrualla2020nerfw}
Ricardo Martin-Brualla, Noha Radwan, Mehdi S.~M. Sajjadi, Jonathan~T. Barron, Alexey Dosovitskiy, and Daniel Duckworth.
\newblock {NeRF in the Wild: Neural Radiance Fields for Unconstrained Photo Collections}.
\newblock In {\em CVPR}, 2021.

\bibitem{barron2021mipnerf}
Jonathan~T. Barron, Ben Mildenhall, Matthew Tancik, Peter Hedman, Ricardo Martin-Brualla, and Pratul~P. Srinivasan.
\newblock Mip-nerf: A multiscale representation for anti-aliasing neural radiance fields, 2021.

\bibitem{muller2022instant}
Thomas M{\"u}ller, Alex Evans, Christoph Schied, and Alexander Keller.
\newblock Instant neural graphics primitives with a multiresolution hash encoding.
\newblock {\em ACM Transactions on Graphics (ToG)}, 41(4):1--15, 2022.

\bibitem{nerfstudio2023}
Matthew Tancik, Ethan Weber, Evonne Ng, Ruilong Li, Brent Yi, Terrance Wang, Alexander Kristoffersen, Jake Austin, Kamyar Salahi, Abhik Ahuja, David Mcallister, Justin Kerr, and Angjoo Kanazawa.
\newblock Nerfstudio: A modular framework for neural radiance field development.
\newblock In {\em Special Interest Group on Computer Graphics and Interactive Techniques Conference Conference Proceedings}. {ACM}, jul 2023.

\bibitem{lin2021barf}
Chen-Hsuan Lin, Wei-Chiu Ma, Antonio Torralba, and Simon Lucey.
\newblock Barf: Bundle-adjusting neural radiance fields.
\newblock In {\em Proceedings of the IEEE/CVF International Conference on Computer Vision}, pages 5741--5751, 2021.

\bibitem{jeong2021self}
Yoonwoo Jeong, Seokjun Ahn, Christopher Choy, Anima Anandkumar, Minsu Cho, and Jaesik Park.
\newblock Self-calibrating neural radiance fields.
\newblock In {\em Proceedings of the IEEE/CVF International Conference on Computer Vision}, pages 5846--5854, 2021.

\bibitem{yen2020inerf}
Lin Yen-Chen, Pete Florence, Jonathan~T. Barron, Alberto Rodriguez, Phillip Isola, and Tsung-Yi Lin.
\newblock {iNeRF}: Inverting neural radiance fields for pose estimation.
\newblock In {\em IEEE/RSJ International Conference on Intelligent Robots and Systems ({IROS})}, 2021.

\bibitem{kendall2015posenet}
Alex Kendall, Matthew Grimes, and Roberto Cipolla.
\newblock Posenet: A convolutional network for real-time 6-dof camera relocalization.
\newblock In {\em Proceedings of the IEEE International Conference on Computer Vision (ICCV)}, December 2015.

\bibitem{piasco2018survey}
Nathan Piasco, D{\'e}sir{\'e} Sidib{\'e}, C{\'e}dric Demonceaux, and Val{\'e}rie Gouet-Brunet.
\newblock A survey on visual-based localization: On the benefit of heterogeneous data.
\newblock {\em Pattern Recognition}, 74:90--109, 2018.

\bibitem{sarlin2023orienternet}
Paul-Edouard Sarlin, Daniel DeTone, Tsun-Yi Yang, Armen Avetisyan, Julian Straub, Tomasz Malisiewicz, Samuel~Rota Bul{\`o}, Richard Newcombe, Peter Kontschieder, and Vasileios Balntas.
\newblock Orienternet: Visual localization in 2d public maps with neural matching.
\newblock In {\em Proceedings of the IEEE/CVF Conference on Computer Vision and Pattern Recognition}, pages 21632--21642, 2023.

\bibitem{schonberger2016structure}
Johannes Schönberger and Jan-Michael Frahm.
\newblock Structure-from-motion revisited.
\newblock 06 2016.

\bibitem{sarlin2020superglue}
Paul-Edouard Sarlin, Daniel DeTone, Tomasz Malisiewicz, and Andrew Rabinovich.
\newblock {SuperGlue}: Learning feature matching with graph neural networks.
\newblock In {\em CVPR}, 2020.

\bibitem{detone2018superpoint}
Daniel DeTone, Tomasz Malisiewicz, and Andrew Rabinovich.
\newblock Superpoint: Self-supervised interest point detection and description.
\newblock In {\em Proceedings of the IEEE conference on computer vision and pattern recognition workshops}, pages 224--236, 2018.

\bibitem{tyszkiewicz2020disk}
Michał~J. Tyszkiewicz, Pascal Fua, and Eduard Trulls.
\newblock Disk: Learning local features with policy gradient, 2020.

\bibitem{sattler2018benchmarking}
Torsten Sattler, Will Maddern, Carl Toft, Akihiko Torii, Lars Hammarstrand, Erik Stenborg, Daniel Safari, Masatoshi Okutomi, Marc Pollefeys, Josef Sivic, et~al.
\newblock Benchmarking 6dof outdoor visual localization in changing conditions.
\newblock In {\em Proceedings of the IEEE conference on computer vision and pattern recognition}, pages 8601--8610, 2018.

\bibitem{walch17spatiallstms}
Florian Walch, Caner Hazirbas, Laura Leal{-}Taix{\'{e}}, Torsten Sattler, Sebastian Hilsenbeck, and Daniel Cremers.
\newblock Image-based localization using lstms for structured feature correlation.
\newblock In {\em IEEE International Conference on Computer Vision (ICCV)}, October 2017.

\bibitem{sucar2021imap}
Edgar Sucar, Shikun Liu, Joseph Ortiz, and Andrew~J Davison.
\newblock imap: Implicit mapping and positioning in real-time.
\newblock In {\em Proceedings of the IEEE/CVF International Conference on Computer Vision}, pages 6229--6238, 2021.

\bibitem{zhu2022nice}
Zihan Zhu, Songyou Peng, Viktor Larsson, Weiwei Xu, Hujun Bao, Zhaopeng Cui, Martin~R Oswald, and Marc Pollefeys.
\newblock Nice-slam: Neural implicit scalable encoding for slam.
\newblock In {\em Proceedings of the IEEE/CVF Conference on Computer Vision and Pattern Recognition}, pages 12786--12796, 2022.

\bibitem{zhu2023nicer}
Zihan Zhu, Songyou Peng, Viktor Larsson, Zhaopeng Cui, Martin~R Oswald, Andreas Geiger, and Marc Pollefeys.
\newblock Nicer-slam: Neural implicit scene encoding for rgb slam.
\newblock {\em arXiv preprint arXiv:2302.03594}, 2023.

\bibitem{maggio2023loc}
Dominic Maggio, Marcus Abate, Jingnan Shi, Courtney Mario, and Luca Carlone.
\newblock Loc-nerf: Monte carlo localization using neural radiance fields.
\newblock In {\em 2023 IEEE International Conference on Robotics and Automation (ICRA)}, pages 4018--4025. IEEE, 2023.

\bibitem{liu2023nerfloc}
Jianlin Liu, Qiang Nie, Yong Liu, and Chengjie Wang.
\newblock Nerf-loc: Visual localization with conditional neural radiance field, 2023.

\bibitem{hou2022implicit}
Yuxin Hou, Tianwei Shen, Tsun-Yi Yang, Daniel DeTone, Hyo~Jin Kim, Chris Sweeney, and Richard Newcombe.
\newblock Implicit map augmentation for relocalization.
\newblock In {\em European Conference on Computer Vision}, pages 621--638. Springer, 2022.

\bibitem{he2015deep}
Kaiming He, Xiangyu Zhang, Shaoqing Ren, and Jian Sun.
\newblock Deep residual learning for image recognition, 2015.

\bibitem{DenDon09Imagenet}
Jia Deng, Wei Dong, Richard Socher, Li-Jia Li, Kai Li, and Li~Fei-Fei.
\newblock Imagenet: A large-scale hierarchical image database.
\newblock In {\em Computer Vision and Pattern Recognition, 2009. CVPR 2009. IEEE Conference on}, pages 248--255. IEEE, 2009.

\bibitem{glocker2013real}
Ben Glocker, Shahram Izadi, Jamie Shotton, and Antonio Criminisi.
\newblock Real-time rgb-d camera relocalization.
\newblock In {\em 2013 IEEE International Symposium on Mixed and Augmented Reality (ISMAR)}, pages 173--179. IEEE, 2013.

\bibitem{polycam2022}
Polycam.
\newblock \url{https://poly.cam/}, 2022.
\newblock Accessed on 5 October 2022.

\bibitem{lindenberger2023lightglue}
Philipp Lindenberger, Paul-Edouard Sarlin, and Marc Pollefeys.
\newblock {LightGlue: Local Feature Matching at Light Speed}.
\newblock In {\em ICCV}, 2023.

\bibitem{lin2023parallel}
Yunzhi Lin, Thomas M{\"u}ller, Jonathan Tremblay, Bowen Wen, Stephen Tyree, Alex Evans, Patricio~A Vela, and Stan Birchfield.
\newblock Parallel inversion of neural radiance fields for robust pose estimation.
\newblock In {\em 2023 IEEE International Conference on Robotics and Automation (ICRA)}, pages 9377--9384. IEEE, 2023.

\end{thebibliography}
}

\end{document}